# MetaOpenFOAM 2.0: Large Language Model Driven Chain of Thought for Automating CFD Simulation and Post-Processing

Yuxuan Chen[a], Xu Zhu[a], Hua Zhou[a], Zhuyin Ren[a*]

[a] *Institute for Aero Engine, Tsinghua University, Beijing 100084, China*
*\* Corresponding author:* zhuyinren@tsinghua.edu.cn

**Abstract**

Computational Fluid Dynamics (CFD) is widely used in aerospace, energy, and biology to model fluid flow, heat transfer, and chemical reactions. While Large Language Models (LLMs) have transformed various domains, their application in CFD remains limited, particularly for complex tasks like post-processing. To bridge this gap, we introduce MetaOpenFOAM 2.0, which leverages Chain of Thought (COT) decomposition and iterative verification to enhance accessibility for non-expert users through natural language inputs. Tested on a new benchmark covering simulation (fluid flow, heat transfer, combustion) and post-processing (extraction, visualization), MetaOpenFOAM 2.0 achieved an Executability score of 6.3/7 and a pass rate of 86.9%, significantly outperforming MetaOpenFOAM 1.0 (2.1/7, 0%). Additionally, it proved cost-efficient, averaging $0.15 per case. An ablation study confirmed that COT-driven decomposition and iterative refinement substantially improved task performance. Furthermore, scaling laws showed that increasing COT steps enhanced accuracy while raising token usage, aligning with LLM post-training scaling trends. These results highlight the transformative potential of LLMs in automating CFD workflows for industrial and research applications. Code is available at https://github.com/Terry-cyx/MetaOpenFOAM

## 1 Introduction

Computational Fluid Dynamics (CFD) is a computational technique that utilizes numerical methods and physical models to solve fluid flow, heat transfer, chemical reactions, and other related processes (Blazek, 2015). It is widely applied in fields such as aerospace, automotive, energy, and biology (An et al., 2020; Mao et al., 2023; Wang et al., 2011; Wei et al., 2023). Current industrial CFD software, such as Fluent (Manual, 2009) and COMSOL (Multiphysics, 1998), typically rely on a GUI interface where users manually select appropriate numerical methods and physical models to perform simulations. However, with the increasing integration of models across various fields, the complexity of these software interfaces has grown significantly. Non-expert users may struggle to make informed choices regarding model selection, leading to a lack of user-friendliness. Meanwhile, significant amounts of CFD simulation data have already been generated in industry by experts for specific cases. Leveraging these case libraries could help non-expert users quickly perform relevant CFD simulations. Recently, with the advancement of Natural Language Processing (NLP) technologies, especially the emergence of Large Language Models (LLMs) (Akata et al., 2023; Du et al., 2023; Hong et al., 2023; Wang et al., 2024; Wang et al., 2023; Zhuge et al., 2023), unprecedented potential has been demonstrated in making CFD simulation software more accessible. These models have made it possible to apply CFD case libraries and develop LLM-driven CFD simulation software using natural language as input.

Since the discovery of pre-trained scaling laws (Kaplan et al., 2020), which describe the relationship between model size, dataset size, and performance, LLMs have undergone rapid development, resulting in the emergence of remarkable models such as GPT-4o (OpenAI, 2024a), Llama 3.1 (AI, 2024), and Qwen 2.5 (Yang et al., 2024). However, as the development of pre-trained scaling laws reached a bottleneck (Kumar et al., 2024), OpenAI introduced the GPT-o1 model (OpenAI, 2024b), signaling the discovery of post-trained scaling laws. The GPT-o1 model integrates process rewards through the application of Chain of Thought (COT) and Reinforcement Learning (RL) techniques, enabling it to achieve groundbreaking results in addressing slow-thinking or complex problems. As a result, enhancing LLM capabilities using COT has become the dominant approach for post-training (Wei et al., 2022). Numerous forms of COT have been proposed recently, including problem decomposition (QDCOT) (Zhou et al., 2022), iterative verification and refinement (ICOT) (Paul et al., 2024), data augmentation (Diao et al., 2023), and sampling (Wang et al., 2022). Due to the complexity, high error rates, and specialized nature of CFD tasks, it is necessary to develop a specialized COT that integrates problem decomposition, iterative verification and refinement, and data augmentation.

MetaOpenFOAM 1.0 (Chen et al., 2024), as a novel CFD simulation software based on natural language input, has been tested on a wide range of multidimensional flow problems, encompassing both compressible and incompressible flows with various physical processes such as turbulence, heat transfer, and combustion. MetaOpenFOAM 1.0 have achieved a high pass rate in these cases (85%), with each test case costing an average of only $0.22. However, these tests excluded CFD post-processing, a crucial part of the workflow. Despite the success of MetaOpenFOAM 1.0, its framework still leaves considerable room for improvement. For example, MetaOpenFOAM 1.0, in terms of its framework, does not utilize multiple QDCOT and ICOT for CFD simulation and post-processing. Instead, it relies on overall decomposition with a single QDCOT and overall verification with a single ICOT, which may result in suboptimal performance when handling complex CFD simulation and post-processing tasks. This is analogous to the concept in Reinforcement Learning (RL) for LLMs, where process rewards are more effective than overall rewards when dealing with slow-thinking or complex problems. Therefore, this paper proposes a new version, MetaOpenFOAM 2.0, which leverages multiple QDCOT and ICOT to assist the LLM in automating both CFD simulation and post-processing. It aims to discover a scaling law for COT that supports the rationale behind a general CFD simulation workflow with COT.

Our contributions can be summarized as follows:

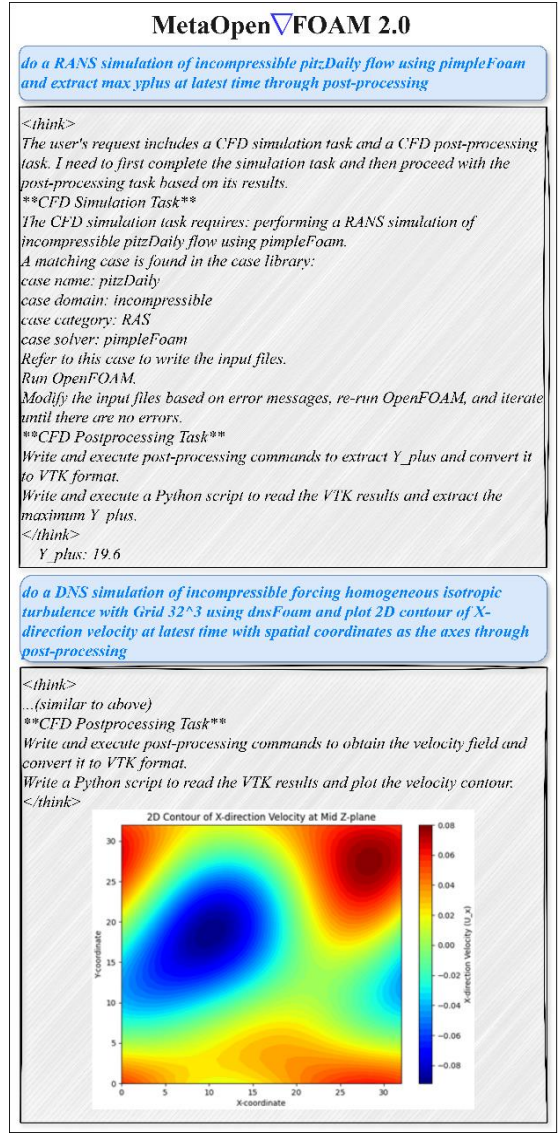

Figure 1: Capabilities of MetaOpenFOAM 2.0 for different CFD simulation and postprocessing task.

- We introduce **MetaOpenFOAM 2.0**, an LLM-driven framework that automates CFD simulations and post-processing through COT, improving accessibility for non-experts.
- Benchmark tests show MetaOpenFOAM 2.0 significantly outperforms its predecessor, achieving an executability score of 6.3 (vs. 2.1 in MetaOpenFOAM 1.0) with a pass rate of 86.9% and an average cost of just $0.15 per case.
- **Scaling law** analysis confirms that increasing decomposition and verification steps enhances executability while raising token usage, aligning with LLM post-training trends.

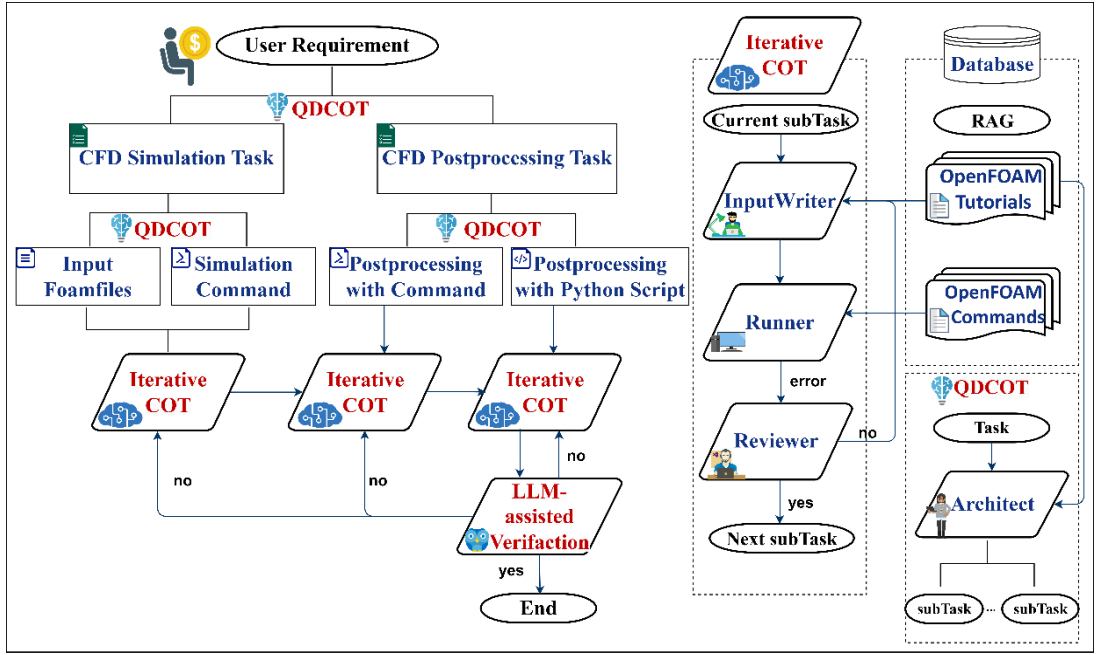

Figure 2: Framework of MetaOpenFOAM 2.0. Where QDCOT means chain of thought (COT) with question decomposition, Iterative COT (ICOT) means COT with iterative verification and refinement, and RAG means retrieval-augmented generation.

## 2 Methodology

### 2.1 MetaOpenFOAM 2.0 Framework

Figure 1 illustrates how MetaOpenFOAM 2.0 addresses complex CFD simulation and post-processing tasks using COT with question decomposition (QDCOT) and iterative verification and refinement (Iterative COT or ICOT).

First, MetaOpenFOAM 2.0 employs two sequential QDCOT steps, each facilitated by an Architect agent to decompose tasks. Initially, the user requirements are divided into two primary tasks: CFD simulation and CFD post-processing task. Each primary task is further broken down into executable subtasks, including writing CFD input files, executing simulation Linux commands, executing post-processing Linux commands, and executing post-processing Python scripts.

For each subtask, an ICOT process is applied. Within each ICOT, several specialized agents collaborate: the InputWriter agent writes and rewrites the input files, Linux commands, or Python scripts required for the subtask; the Runner agent executes the relevant Linux commands; and the Reviewer agent examines errors reported by the Runner and provides feedback to the InputWriter for further refinements. Once a subtask passes the Reviewer's checks without errors, the process moves on to the next subtask, continuing ICOT until all subtasks are completed.

After all subtasks are executed, an additional LLM-assisted verification step is performed. This step evaluates the outputs against various criteria, including user requirements, physical accuracy, flow characteristics, numerical accuracy and boundary condition consistency. If the results are deemed unsatisfactory, the verification process identifies which subtask requires correction. ICOT is then resumed from that specific subtask, iterating until the LLM-assisted verification is successful.

In MetaOpenFOAM 1.0 (Chen et al., 2024), the Architect, InputWriter, Runner, and Reviewer agents were organized into a single QDCOT followed by a single ICOT. The framework lacked LLM-assisted verification and had limited functionalities, focusing solely on decomposing input files, writing/rewriting input files, running solver and reviewing errors. Compared to MetaOpenFOAM 1.0, MetaOpenFOAM 2.0 framework offers a more advanced and versatile set of COT methodologies and agent functionalities, enabling it to handle a broader range of CFD simulation and post-processing tasks more effectively. The RAG technology from MetaOpenFOAM 1.0, which integrates a searchable database of OpenFOAM documentation to enhance agents' task performance with minimal user intervention, is also retained in MetaOpenFOAM 2.0.

The detailed algorithm of the general CFD simulation workflow with COT is presented in **Appendix A.1**.

## 3 Experiment

### 3.1 Setup

MetaGPT v0.8.0 (Hong et al., 2023) was chosen for the integration of different agents, while OpenFOAM 10 (Jasak et al., 2007) was employed for CFD simulations due to its stability and dependability as an open-source solver. GPT-4o (OpenAI, 2024a) was selected as the primary LLM, owing to its exceptional performance. To minimize randomness in the generated output, the temperature parameter, which controls the degree of randomness in LLM-generated text, was configured to 0.01, ensuring more deterministic and focused results. The impact of temperature settings on model performance is discussed in (Chen et al., 2024).

In terms of RAG, LangChain v0.1.19 (Chase, 2022) facilitated the connection between the agents and the database. The FAISS vector store (Douze et al., 2024), recognized for its high efficiency and ease of use, was utilized as the vector store for the database, and OpenAIEmbeddings were chosen for embedding the data segments. The "similarity" approach was employed to identify and match related chunks of data. The simplest stacking approach was used, combining retrieved documents with user input messages.

### 3.2 Benchmarking Natural Language Input for CFD Simulation and Postprocessing

The only public benchmark for CFD simulations, released with MetaOpenFOAM 1.0 (Chen et al., 2024), lacks post-processing tasks. This study introduces a new benchmark covering both CFD simulation and post-processing.

Post-processing tasks are categorized into visualization (e.g., contour plots, line graphs) and extraction (e.g., computing averages, extrema). The benchmark includes 13 user requirements, with six focusing on PitzDaily post-processing (e.g., extracting max y+, Courant number, velocity contours) and seven covering post-processing for different simulations, including fluid flow (HIT), heat transfer (BuoyantCavity), and combustion (CounterFlowFlame). These cases, adapted from OpenFOAM tutorials, include customized parameters and tasks. Detailed user requirements are provided in **Appendix A.2**.

### 3.3 Evaluation Metrics of MetaOpenFOAM 2.0

Currently, the evaluation metrics in MetaOpenFOAM 1.0 (Chen et al., 2024) are only applicable to cases where the user requirement is a CFD simulation task and do not account for post-processing tasks or the overall evaluation based on LLM. Therefore, for the newly established benchmark that includes both CFD simulation and post-processing tasks, it is necessary to develop new evaluation metrics to assess the performance of MetaOpenFOAM 2.0. For CFD software with natural language input, the performance can be evaluated using the following three metrics: the first two metrics, A and B, are used to evaluate single experiments, while the third metric, C, is used to assess multiple experiments.

Single experiments can be evaluated by the following metrics:

**(A) Executability**:

This metric evaluates the results of MetaOpenFOAM 2.0 on a scale from 0 (failure) to 7 (flawless). Scores from 0 to 3 correspond to the evaluation of simulation tasks, while scores from 4 to 5 correspond to the evaluation of post-processing tasks. A score of 6 represents the evaluation of overall results by the LLM, and a score of 7 involves human judgment. Specifically: A score of '0' indicates grid generation failure; '1' indicates grid generation success but running failure; '2' indicates that the case is runnable but does not converge; '3' indicates that the case runs to the endTime specified in the controlDict; A score of '4' indicates that the post-processing executable command runs successfully and generates the post-processing files; A score of '5' indicates that the Python script successfully reads the CFD simulation/post-processing result files and completes the post-processing tasks; A score of '6' means the LLM reviews the results generated by the Python script (e.g., contours, extracted values) and assesses their validity based on factors such as user requirements, physical accuracy, flow characteristics, numerical accuracy and boundary condition consistency; A score of '7' indicates flawless results, requiring human judgment to verify whether the results are physically accurate and meet all user requirements.

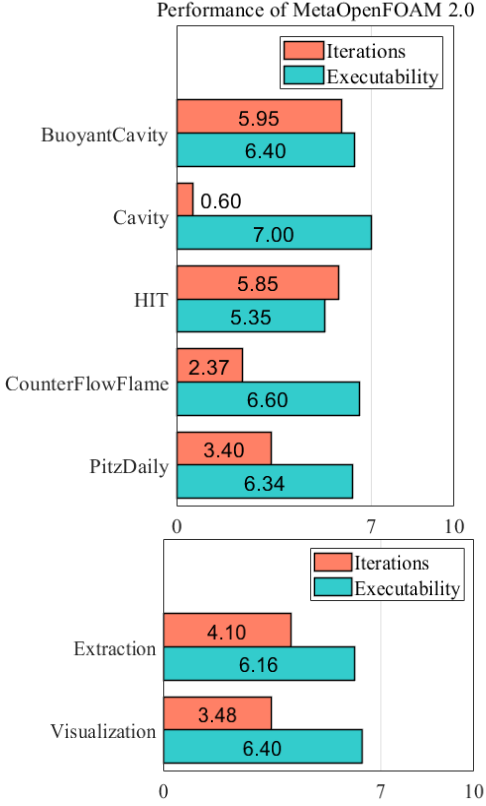

Figure 3: Performance of MetaOpenFOAM 2.0 on Executability and Iterations for different simulation tasks and postprocessing tasks.

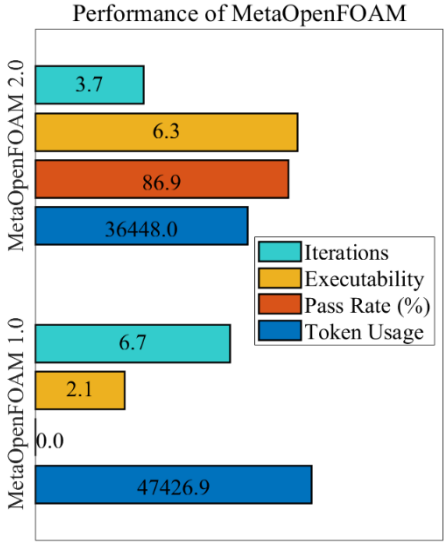

Figure 4: Performance of MetaOpenFOAM 1.0 (Chen et al., 2024) and MetaOpenFOAM 2.0 on the benchmark.

**(B) Cost**:

The cost evaluation includes the following components: (1) the number of iterations, which covers iterations for simulation, post-processing with command, and post-processing with Python script; (2) token usage, which is divided into prompt tokens and completion tokens based on LLM input and output, and further categorized into tokens for iterations and tokens for non-iterations based on the iterative and non-iterative processes in MetaOpenFOAM 2.0; and (3) expenses, which are proportional to token usage.

For multiple experiments, a new metric needs to be added:

**(C) Pass@k**:

This metric quantifies the probability that at least one of the k generated input file samples successfully passes the tests. It evaluates the model's ability to produce correct results within k attempts. To assess pass@k for MetaOpenFOAM, we adopt the unbiased version of pass@k as described in (Chen et al., 2021; Dong et al., 2024):

$$pass@k := \mathop{E}_{problems}\left[1 - \frac{\binom{n-c}{k}}{\binom{n}{k}}\right],$$

where $n$ refers to the total number of input samples generated for each user requirement, and $c$ denotes the number of these samples that pass the test, i.e., achieve an executability score of 7. For pass@k evaluation, we generate $n \geqslant k$ samples per task (with $n = 10$ and $k = 1$ in this study), count the number of correct samples $c \leqslant n$ which pass tests, and compute the unbiased estimate.

### 3.4 Main Results

MetaOpenFOAM 2.0 demonstrates strong performance across 13 CFD simulation and post-processing tasks, achieving an average Executability score of 6.3 and a Pass@1 rate of 86.9%. To reduce variability, all results were averaged over 10 independent runs. In terms of efficiency, the model consumes an average of 36,448 tokens per test case, with an estimated cost of $0.15 per case—significantly lower than hiring domain experts. These results underscore its potential to lower entry barriers, reduce labor costs, and enhance the efficiency of CFD case setup for complex tasks. Detailed results can be found in **Appendix A.3**.

Figure 3 (up) and Figure 3 (down) illustrate the Executability and Iterations for various simulation and post-processing tasks, respectively. Among the simulation tasks, Cavity achieved the highest Executability score of 7.0, while HIT had the lowest at 5.35, with other tasks ranging between 6.3 and 6.6. HIT's lower Executability is likely due to a low-similarity match in the case database (see

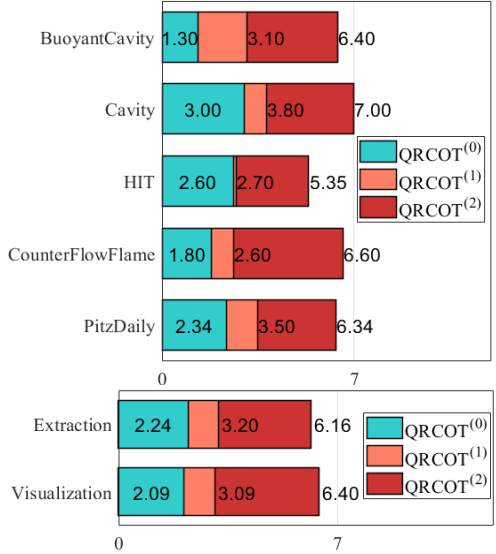

Figure 5: Executability of MetaOpenFOAM with different removals in different simulation tasks (up) and different postprocessing tasks (down). Where QRCOT$^{(0)}$ means removing postprocessing with python script and postprocessing with command, QRCOT$^{(1)}$ means removing postprocessing with python script, QRCOT$^{(2)}$ means removing nothing, i.e., the complete and original MetaOpenFOAM 2.0.

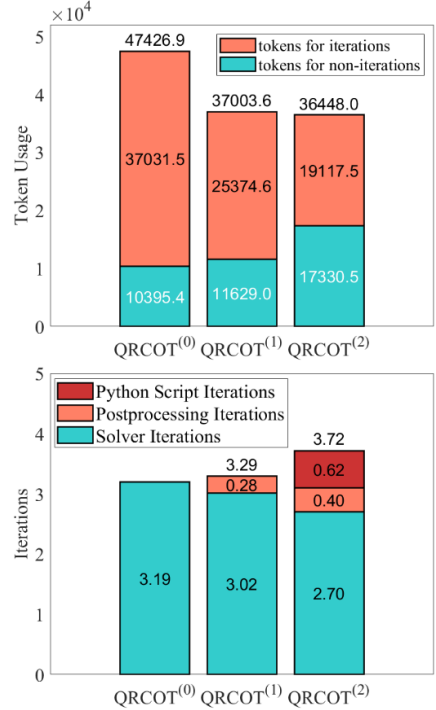

Figure 6: Token usage (up) and Iterations (down) of MetaOpenFOAM with different removals. Token usage for iterations (including review and rewrite) and non-iterations (including first write input files of OpenFOAM and Linux commands) Iterations for Simulation (Iterations of simulation of CFD solver), Postprocessing (Iterations of postprocessing with Linux commands), and Python Script (Iterations of postprocessing with python script).

**Appendix A.4**). Iteration counts did not exhibit a strong negative correlation with Executability; for instance, BuoyantCavity required more Iterations than HIT and PitzDaily despite its higher Executability. However, tasks with lower Executability generally reached the maximum Iterations (set to 10). For post-processing tasks, Visualization (e.g., contour and profile plotting) achieved slightly higher Executability and required fewer Iterations than Extraction (e.g., computing averages, maxima, and minima). This difference may stem from the varying number of cases across categories. Overall, MetaOpenFOAM 2.0 demonstrated robust performance across both extraction and visualization tasks, highlighting its versatility in CFD post-processing.

Figure 4 presents a performance comparison between MetaOpenFOAM 1.0 (Chen et al., 2024) and MetaOpenFOAM 2.0, including metrics such as Iterations, Executability, Pass rate (Pass@1), and token usage. As shown, MetaOpenFOAM 2.0 outperforms MetaOpenFOAM 1.0 in terms of both Executability and Pass@1 on the new benchmark. Furthermore, in terms of cost, MetaOpenFOAM 2.0 shows a clear reduction in both the number of Iterations and token usage compared to MetaOpenFOAM 1.0. It is evident that MetaOpenFOAM 1.0 struggles to meet the corresponding user requirements when postprocessing tasks are involved, thereby highlighting the necessity of introducing MetaOpenFOAM 2.0.

## 4 Discussion

This section conducts an ablation analysis to evaluate the necessity of the new modules introduced in MetaOpenFOAM 2.0. (The role of the LLM-assisted verification module is explained in **Appendix A.5**) Subsequently, based on the results of the ablation study, a scaling law is found to describe the relationship between token usage and Executability.

### 4.1 Ablation Analysis

This subsection evaluates COT's significance from two distinct perspectives: QRCOT and ICOT. By selectively removing specific modules, the analysis demonstrates the contribution of each approach to the overall performance.

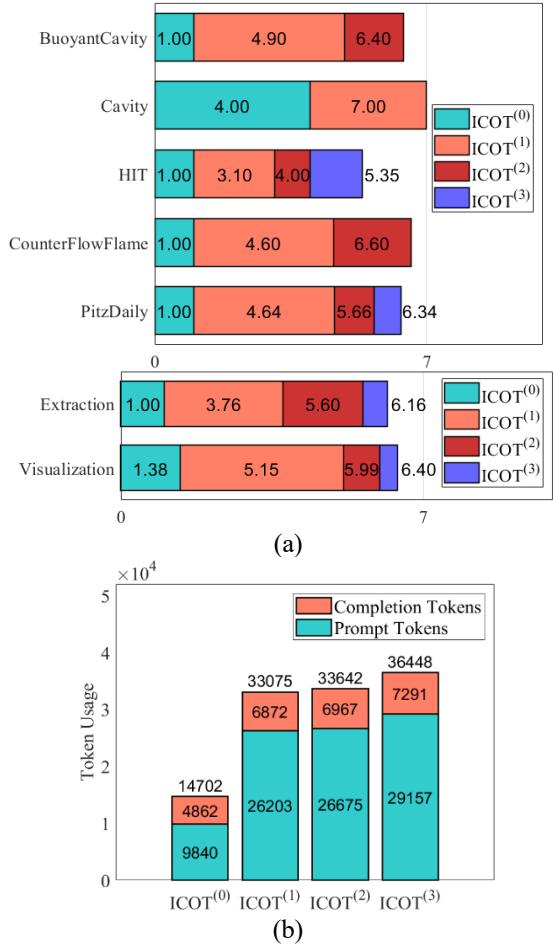

Figure 7: Executability and token usage of MetaOpenFOAM with different removals. Executability with different CFD simulation task (b) Executability with different CFD postprocessing task. (c) Token usage for prompt (input) and completion (output). Where ICOT$^{(i)}$ refers to i iterative chains of thought (ICOT). ICOT $^{(0)}$ means removing reviewer of postprocessing with python script, postprocessing with command and simulation, ICOT $^{(1)}$ means removing reviewer of postprocessing with python script and postprocessing with command, ICOT $^{(2)}$ means removing reviewer of postprocessing with python script, ICOT $^{(3)}$ means removing nothing, i.e., the complete and original MetaOpenFOAM 2.0.

*Ablation of COT with Question Decomposition*

Unlike MetaOpenFOAM 1.0, which handled writing, rewriting, and executing the Allrun script in a single-stage process, MetaOpenFOAM 2.0 introduces a modular workflow with three sequential submodules: simulation execution, command-based postprocessing, and Python-based postprocessing. This decomposition enhances performance and flexibility by addressing the limitations of a monolithic approach.

To assess the necessity of these submodules, two ablation scenarios were analyzed:

- **QRCOT$^{(0)}$**: removing all postprocessing modules, reverting to MetaOpenFOAM 1.0 workflow.
- **QRCOT$^{(1)}$**: removing only the Python-based postprocessing module, relying solely on Linux commands.

Figure 5 shows that Executability improves as more decomposition is introduced. Adding command-based postprocessing increased Executability from approximately 2 to 3 by enhancing simulation completion rates, though postprocessing remained limited. Further integrating Python-based postprocessing boosted Executability to approximately 6, significantly improving postprocessing success. The impact varied by task—BuoyantCavity, with its complex requirements, benefited the most, while HIT, with simpler needs, showed minimal improvement. Across postprocessing tasks, decomposition had a uniform effect on Visualization and Extraction.

Figure 6 illustrates Token Usage and Iterations under different scenarios. More detailed decomposition increased prompt token consumption but reduced overall token usage by minimizing costly simulation iterations. Since simulations require reviewing extensive OpenFOAM inputs, fewer iterations significantly cut token costs. Meanwhile, additional postprocessing steps contributed only marginally to token consumption.

In summary, finer task decomposition improves model precision, reduces error propagation, and enhances execution efficiency, leading to higher success rates. While this approach increases token usage due to added context dependency, the COT with question decomposition effectively balances improved Executability with optimized token consumption, validating its efficiency.

*Ablation of COT with Verification and Refinement*

MetaOpenFOAM 2.0 employs iterative verification and refinement through *InputWriter*, *Runner*, and *Reviewer* across three stages: simulation, command-based postprocessing, and Python-based postprocessing. To assess the impact of removing iterative review, three ablation scenarios were tested:

- **ICOT²**: Reviews only simulations and command-based postprocessing, exiting if Python-based postprocessing fails.
- **ICOT¹**: Reviews only simulations, exiting if either postprocessing step fails.

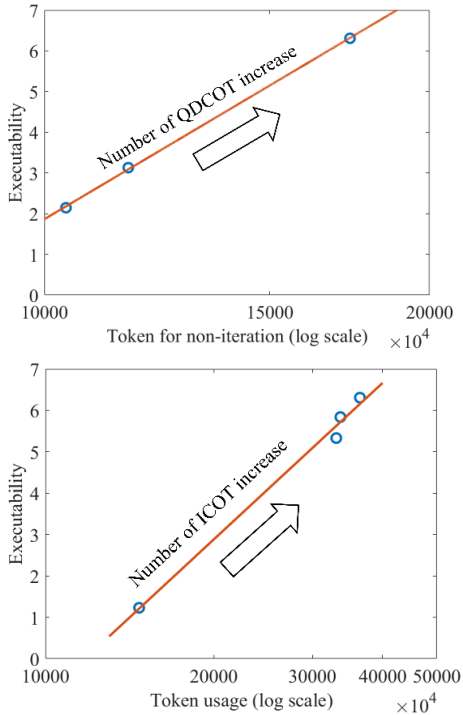

Figure 8: MetaOpenFOAM performance improves smoothly as we increase the number of question decomposed COT (up) iterative COT (down). For optimal performance both two factors must be scaled up in tandem, which also leads to an increase in token usage. The dots represent the results under different COT frameworks, with the lines fitted to these data points.

- **ICOT⁰**: Removes all reviewers, exiting upon any error.

Figure 7 shows that adding reviewers improves Executability but increases token usage due to more iterations. While review benefits vary by task—Cavity saw no improvement—command-based reviews notably enhanced extraction tasks, which are more error-prone than visualization. Token usage analysis reveals that with more reviewers, prompt token consumption rose from 67% to 80%, as additional context from error messages and prior inputs was incorporated into prompts.

### 4.2 Scaling laws in MetaOpenFOAM

During the ablation study, we observed that Executability increases as the number of QDCOT and ICOT increases, aligning with the post-trained scaling law proposed by OpenAI o1 (OpenAI, 2024b).

Based on the results in Section 4.1, there is a clear positive correlation between the number of COT with question decomposition and both Executability and token usage for non-iterations. Figure 8 (up) illustrates this trend, showing that token usage for non-iterations and Executability follow a scaling law similar to the post-trained scaling law proposed by OpenAI o1 (OpenAI, 2024b). This suggests that allocating more tokens for task decomposition improves framework's performance. Similarly, Figure 8 (down) shows that increasing verification and refinement steps follows a scaling law, leading to higher Executability and token usage. This suggests that allocating more tokens for these processes further enhances MetaOpenFOAM's performance.

Based on these two scaling laws, it can be anticipated that for other tasks in the complete CFD simulation workflow, such as mesh generation, pre-processing, and initialization, using QDCOT combined with ICOT to further decompose and iteratively verify and correct these tasks will enhance MetaOpenFOAM's executability across a broader range of test cases.

## 5 Conclusion

MetaOpenFOAM 2.0 enhances the automation of CFD simulations and post-processing through an LLM-driven multi-agent system. Building on MetaOpenFOAM 1.0, it integrates COT with problem decomposition and iterative verification and refinement, making complex CFD tasks more accessible via natural language.

MetaOpenFOAM 2.0 was evaluated on benchmark tasks, including fluid flow, heat transfer, combustion, and post-processing (visualization and extraction), achieving an executability score of 6.3/7 and a pass rate of 86.9%, far surpassing MetaOpenFOAM 1.0 (2.1/7, 0%). With an average cost of $0.15 per case, it offers a highly cost-effective alternative to domain experts.

The ablation study confirmed that COT-based decomposition and iterative verification significantly improve executability. While decomposition reduces iteration-related token usage, it increases non-iterative token consumption. Iterative verification further enhances task success by refining outputs at the cost of more tokens.

Scaling law analysis revealed that increasing decomposition and verification steps improves executability while raising token usage, aligning with LLM post-training trends. These findings highlight MetaOpenFOAM 2.0's potential in automating CFD workflows, providing an efficient and accessible solution for industrial and research applications.

## Acknowledgments

This work was supported by the National Natural Science Foundation of China (No. 52025062 and 52106166). The authors also acknowledge High-Performance Computing Centre at Tsinghua University for providing computational resource. During the preparation of this work the author(s) used ChatGPT in order to improve language and readability. After using this tool/service, the author(s) reviewed and edited the content as needed and take(s) full responsibility for the content of the publication.

## References

AI, M. (2024). *Introducing Llama 3.1: Our most capable models to date*. https://ai.meta.com/blog/meta-llama-3-1/

Akata, E., Schulz, L., Coda-Forno, J., Oh, S. J., Bethge, M., & Schulz, E. (2023). Playing repeated games with large language models. *arXiv preprint arXiv:2305.16867*.

An, J., Wang, H., Liu, B., Luo, K. H., Qin, F., & He, G. Q. (2020). A deep learning framework for hydrogen-fueled turbulent combustion simulation. *International Journal of Hydrogen Energy*, *45*(35), 17992-18000.

Blazek, J. (2015). *Computational fluid dynamics: principles and applications*. Butterworth-Heinemann.

Chase, H. (2022). *LangChain, https://github.com/langchain-ai/langchain*.

Chen, M., Tworek, J., Jun, H., Yuan, Q., Pinto, H. P. D. O., Kaplan, J., Edwards, H., Burda, Y., Joseph, N., & Brockman, G. (2021). Evaluating large language models trained on code. *arXiv preprint arXiv:2107.03374*.

Chen, Y., Zhu, X., Zhou, H., & Ren, Z. (2024). MetaOpenFOAM: an LLM-based multi-agent framework for CFD. *arXiv preprint arXiv:2407.21320*.

Diao, S., Wang, P., Lin, Y., Pan, R., Liu, X., & Zhang, T. (2023). Active prompting with chain-of-thought for large language models. *arXiv preprint arXiv:2302.12246*.

Dong, Y., Jiang, X., Jin, Z., & Li, G. (2024). Self-collaboration code generation via chatgpt. *ACM Transactions on Software Engineering and Methodology*, *33*(7), 1-38.

Douze, M., Guzhva, A., Deng, C., Johnson, J., Szilvasy, G., Mazaré, P.-E., Lomeli, M., Hosseini, L., & Jégou, H. (2024). The faiss library. *arXiv preprint arXiv:2401.08281*.

Du, Y., Li, S., Torralba, A., Tenenbaum, J. B., & Mordatch, I. (2023). Improving factuality and reasoning in language models through multiagent debate. *arXiv preprint arXiv:2305.14325*.

Hong, S., Zheng, X., Chen, J., Cheng, Y., Wang, J., Zhang, C., Wang, Z., Yau, S. K. S., Lin, Z., & Zhou, L. (2023). Metagpt: Meta programming for multi-agent collaborative framework. *arXiv preprint arXiv:2308.00352*.

Jasak, H., Jemcov, A., & Tukovic, Z. (2007). OpenFOAM: A C++ library for complex physics simulations. International workshop on coupled methods in numerical dynamics,

Kaplan, J., McCandlish, S., Henighan, T., Brown, T. B., Chess, B., Child, R., Gray, S., Radford, A., Wu, J., & Amodei, D. (2020). Scaling laws for neural language models. *arXiv preprint arXiv:2001.08361*.

Kumar, T., Ankner, Z., Spector, B. F., Bordelon, B., Muennighoff, N., Paul, M., Pehlevan, C., Ré, C., & Raghunathan, A. (2024). Scaling laws for precision. *arXiv preprint arXiv:2411.04330*.

Manual, U. (2009). ANSYS FLUENT 12.0. *Theory Guide*, *67*.

Mao, R., Lin, M., Zhang, Y., Zhang, T., Xu, Z.-Q. J., & Chen, Z. X. (2023). DeepFlame: A deep learning empowered open-source platform for reacting flow simulations. *Computer Physics Communications*, *291*, 108842.

Multiphysics, C. (1998). Introduction to comsol multiphysics®. *COMSOL Multiphysics, Burlington, MA, accessed Feb*, *9*(2018), 32.

OpenAI. (2024a). *Hello gpt-4o*. https://openai.com/index/hello-gpt-4o/

OpenAI. (2024b). *Introducing OpenAI o1*. https://openai.com/o1/

Paul, D., West, R., Bosselut, A., & Faltings, B. (2024). Making Reasoning Matter: Measuring and Improving Faithfulness of Chain-of-Thought Reasoning. *arXiv preprint arXiv:2402.13950*.

Wang, L., Ma, C., Feng, X., Zhang, Z., Yang, H., Zhang, J., Chen, Z., Tang, J., Chen, X., & Lin, Y. (2024). A survey on large language model based autonomous agents. *Frontiers of Computer Science*, *18*(6), 186345.

Wang, X., Wei, J., Schuurmans, D., Le, Q., Chi, E., Narang, S., Chowdhery, A., & Zhou, D. (2022). Self-consistency improves chain of thought reasoning in language models. *arXiv preprint arXiv:2203.11171*.

Wang, Y., Chatterjee, P., & de Ris, J. L. (2011). Large eddy simulation of fire plumes. *Proceedings of the Combustion Institute*, *33*(2), 2473-2480.

Wang, Z., Mao, S., Wu, W., Ge, T., Wei, F., & Ji, H. (2023). Unleashing the emergent cognitive synergy in large language models: A task-solving agent through multi-persona self-collaboration. *arXiv preprint arXiv:2307.05300*.

Wei, J., An, J., Wang, N., Zhang, J., & Ren, Z. (2023). Velocity nonuniformity and wall heat loss coupling effect on supersonic mixing layer flames. *Aerospace Science and Technology*, *141*, 108545.

Wei, J., Wang, X., Schuurmans, D., Bosma, M., Xia, F., Chi, E., Le, Q. V., & Zhou, D. (2022). Chain-of-thought prompting elicits reasoning in large language models. *Advances in neural information processing systems*, *35*, 24824-24837.

Yang, A., Yang, B., Zhang, B., Hui, B., Zheng, B., Yu, B., Li, C., Liu, D., Huang, F., & Wei, H. (2024). Qwen2. 5 Technical Report. *arXiv preprint arXiv:2412.15115*.

Zhou, D., Schärli, N., Hou, L., Wei, J., Scales, N., Wang, X., Schuurmans, D., Cui, C., Bousquet, O., & Le, Q. (2022). Least-to-most prompting enables complex reasoning in large language models. *arXiv preprint arXiv:2205.10625*.

Zhuge, M., Liu, H., Faccio, F., Ashley, D. R., Csordás, R., Gopalakrishnan, A., Hamdi, A., Hammoud, H. A. A. K., Herrmann, V., & Irie, K. (2023). Mindstorms in natural language-based societies of mind. *arXiv preprint arXiv:2305.17066*.

## A Appendices

### A.1 General CFD Simulation Workflow with COT

```
Algorithm 1 CFD Simulation Workflow
Input: User requirement
Output: CFD simulation results (data, figures, etc.)
 1: function SIMULATECFD(user_requirement)
 2:     Decompose user requirement into tasks.
 3:     for each task do
 4:         Decompose task into subtasks corresponding to systems.
 5:     end for
 6:     while subtasks is not empty do
 7:         for each subtask do
 8:             Write input files, execute command.
 9:             Run command.
10:             while return an error do
11:                 Iterations += 1.
12:                 if iterations exceeds max then
13:                     break
14:                 end if
15:                 Review error, rewrite files, re-run.
16:             end while
17:         end for
18:         Validate results using LLM.
19:         if errors detected then
20:             Update subtasks with subtasks_error.
21:         else
22:             Set subtasks to empty.
23:         end if
24:     end while
25:     Return: CFD simulation results.
26: end function
```

Algorithm 1 illustrates the general LLM-driven COT approach for CFD simulations with natural language inputs. This algorithm is not limited to the OpenFOAM CFD software platform, nor is it restricted to CFD simulation and CFD post-processing tasks.

The algorithm begins with a QDCOT step to decompose the user requirements. For CFD, assuming the geometry is already defined, a complete simulation process typically involves grid generation, pre-processing (e.g., calculating the 1D laminar flamelet in the flamelet-generated manifold (FGM) model), initialization, simulation, and post-processing. In MetaOpenFOAM 1.0, this entire process was not decomposed into separate tasks. However, in MetaOpenFOAM 2.0, the process is divided into two main tasks: CFD simulation and CFD post-processing. The former encompasses grid generation, pre-processing, initialization, and simulation, while the latter covers post-processing. In addition to these tasks, there may also be tasks such as sensitivity analysis, parameter calibration, and geometry optimization, which are performed based on the results of the CFD simulation.

The first QDCOT step decomposes the overall process, and once the tasks are identified, a second QDCOT is performed based on the number of systems to be executed, resulting in the identification of subtasks. For example, the post-processing task may be difficult to complete using only Linux commands. Therefore, in MetaOpenFOAM 2.0, this task is divided into two subtasks: Postprocessing with command and Postprocessing with Python script, thus facilitating more efficient task completion. The number of subtasks created depends on the number of systems to be executed.

Next, each subtask undergoes ICOT. It is important to note that the agents within each ICOT must be adapted according to the specific subtask. For instance, in grid generation, the Reviewer agent not only reviews error messages when a failure occurs but also checks the quality parameters of the grid after a successful run. Only when the grid generation meets the required standards can the process proceed to the next subtask. Once all subtasks are completed, and LLM-assisted verification is successfully passed, the CFD simulation based on natural language input is considered complete.

## A.2 Detailed Benchmarking Natural Language Input for CFD Simulation and Postprocessing

①PitzDaily, Max yplus: do a RANS simulation of incompressible pitzDaily flow using pimpleFoam and extract max yplus at latest time through post-processing.

②PitzDaily, Max Co: do a RANS simulation of incompressible PitzDaily flow using pimpleFoam and extract max Courant number at latest time through post-processing.

③PitzDaily, Plot profile of U in X = 0.05 m: do a RANS simulation of incompressible PitzDaily flow using pimpleFoam and plot the X velocity at X = 0.05 m with spatial Y coordinate as the axes at latest time through post-processing.

④PitzDaily, Plot contour of U: do a RANS simulation of incompressible PitzDaily flow using pimpleFoam with inlet velocity = 0.1 m/s, and plot U contour through post-processing.

⑤PitzDaily, Plot contour of U and streamlines: do a RANS simulation of incompressible PitzDaily flow using pimpleFoam with inlet velocity = 0.1 m/s, and then carry out post-processing to generate a 2D contour plot of the velocity distribution in the X-direction, overlaid with streamlines. The X-direction velocity contours should be drawn using a rainbow color map to represent the variation in velocity magnitude, while the streamlines should illustrate the flow direction. The final plot should display spatial coordinates as axes, with both the X-direction velocity contours and streamlines on the same figure, ensuring proper labeling of the colorbar, axis, and units.

⑥CounterFlowFlame Max T: do a 2D laminar simulation of counterflow flame using reactingFoam in combustion with grid 50*20*1 and extract max temperature at latest time through post-processing.

⑦CounterFlowFlam plot contour of T e: do a 2D laminar simulation of counterflow flame using reactingFoam in combustion with grid 50*20*1 and plot 2D contour of temperature distribution at latest time with spatial coordinates as the axes through post-processing.

⑧CounterFlowFlame plot profile of T at Y = 0 m: do a 2D laminar simulation of counterflow flame using reactingFoam in combustion with grid 50*20*1 and plot temperature at Y = 0 m with spatial X coordinate as the axes at latest time through post-processing.

⑨HIT plot X-direction velocity: do a DNS simulation of incompressible forcing homogeneous isotropic turbulence with Grid 32^3 using dnsFoam and plot 2D contour of X-direction velocity at latest time with spatial coordinates as the axes through post-processing.

⑩HIT average TKE: do a DNS simulation of incompressible forcing homogeneous isotropic turbulence with Grid 32^3 using dnsFoam and use the calculated velocity field U at latest time to compute the average turbulent kinetic energy across the entire domain with the formula average(1/2*U^2) through post-processing.

⑪Cavity plot TKE: do a 2D RANS simulation of incompressible cavity flow using pisoFoam, with RANS model: RNGkEpsilon, and plot 2D contours of turbulent kinetic energy distribution with spatial coordinates as the axes through post-processing.

⑫BuoyantCavity plot contour of T: do a RANS simulation of buoyantCavity using buoyantFoam, which investigates natural convection in a heat cavity with a temperature difference of 20K is maintained between the hot and cold; the remaining patches are treated as adiabatic. And the postprocessing task is to plot 2D contour of temperature distribution with spatial coordinates as the axes through post-processing.

⑬BuoyantCavity max X velocity: do a RANS simulation of buoyantCavity using buoyantFoam, which investigates natural convection in a heat cavity with a temperature difference of 20K is maintained between the hot and cold; the remaining patches are treated as adiabatic. And the postprocessing task is to extract the max velocity in X direction through post-processing.

## A.3 Detailed performance and test examples of main results

Table 2 supplements the results presented in Table 1. In this table, *Simulation Iterations* refers to the iterations used for simulation, *Postprocessing Iterations* represents the iterations used for postprocessing with commands, and *Python Script Iterations* indicates the iterations for postprocessing with Python scripts. *Prompt Tokens* refer to the tokens consumed by the input prompts provided to the LLM, while *Completion Tokens* denote the tokens generated in the LLM's output. As shown, the majority of iterations occur during the simulation stage, suggesting that more errors are encountered during the simulation, whereas fewer occur during postprocessing. The ratio of prompt tokens to completion tokens is approximately 4:1, indicating that a significant amount of prompting was used to guide the LLM towards generating better outputs. These prompts include repeated input of previously generated file information and error messages from system execution, which results in a higher token usage than if the LLM output only contained file information. Figure 9 and Table 3 present the postprocessing results for visualization and extraction tasks in MetaOpenFOAM 2.0, with the results randomly selected from those that passed the tests. It can be observed that the post-processing results with an executability score of 7 are generally sufficient to meet engineering post-processing requirements, thereby demonstrating the feasibility of MetaOpenFOAM 2.0 as a CFD simulation software capable of performing both simulation and post-processing tasks using natural language input.

Table 1 Performance of MetaOpenFOAM 2.0. Where the values of Executability, Token Usage, and Iterations are the averages taken from 10 random generations.

| Simulation Task | Postprocessing Task | Executability (Max: 7) | Token Usage | Iterations (Max: 10) | Pass@1 (%) |
|---|---|---|---|---|---|
| *pitzDaily* | *Extract max Yplus* | 6.6 | 26764.6 | 2.9 | 90 |
| *pitzDaily* | *Extract max Courant number* | 6.1 | 26966.5 | 2.9 | 80 |
| *pitzDaily* | *Plot profile of U* | 6.2 | 29598.3 | 3.4 | 80 |
| *pitzDaily* | *Plot contour of U* | 6.7 | 24638.6 | 3.2 | 90 |
| *pitzDaily* | *Plot contour of U with streamlines* | 6.1 | 32397.4 | 4.6 | 80 |
| *CounterFlowFlame* | *Extract max T* | 6.4 | 44450.5 | 3.2 | 90 |
| *CounterFlowFlame* | *Plot contour of T* | 7.0 | 37240.9 | 1.4 | 100 |
| *CounterFlowFlame* | *Plot profile of T* | 6.4 | 41349.0 | 2.5 | 90 |
| *HIT* | *Plot X-velocity* | 5.4 | 29951.0 | 5.9 | 80 |
| *HIT* | *Extract average TKE* | 5.3 | 36039.2 | 5.8 | 70 |
| *Cavity* | *Plot contour of TKE* | 7.0 | 19560.6 | 0.6 | 100 |
| *BuoyantCavity* | *Plot contour of T* | 6.4 | 64452.3 | 6.2 | 90 |
| *BuoyantCavity* | *Extract max X-velocity* | 6.4 | 60414.6 | 5.7 | 90 |
| *Average* | | 6.3 | 36448.0 | 3.7 | 86.9 |

Table 2 Performance of MetaOpenFOAM 2.0. Where the values of Iterations and Token Usage are the averages taken from 10 random generations.

| CFD Task | Simulation Iterations | Postprocessing Iterations | Python Script Iterations | Prompt Tokens | Completion Tokens |
|---|---|---|---|---|---|
| ① | 1.8 | 1.1 | 0 | 21387.2 | 5377.4 |
| ② | 1.4 | 0 | 1.5 | 21342.8 | 5623.7 |
| ③ | 0.8 | 2.4 | 0.2 | 24102.9 | 5495.4 |
| ④ | 1.2 | 0 | 2 | 19240.2 | 5398.4 |
| ⑤ | 3.8 | 0.4 | 0.4 | 26204 | 6193.4 |
| ⑥ | 2.2 | 1 | 0 | 34519.8 | 9931.7 |
| ⑦ | 1.4 | 0 | 0 | 28523.5 | 8717.4 |
| ⑧ | 2.3 | 0 | 0.2 | 32711 | 8638 |
| ⑨ | 3.3 | 0 | 2.6 | 23511.8 | 6439.2 |
| ⑩ | 4.4 | 0.3 | 1.1 | 29077.6 | 6961.6 |
| ⑪ | 0.6 | 0 | 0 | 15083.7 | 4476.9 |
| ⑫ | 6.2 | 0 | 0 | 53815.1 | 10637.2 |

| ⑬ | 5.7 | 0 | 0 | 49526.3 | 10888.3 |
| *Average* | 2.7 | 0.4 | 0.62 | 29157.4 | 7290.7 |

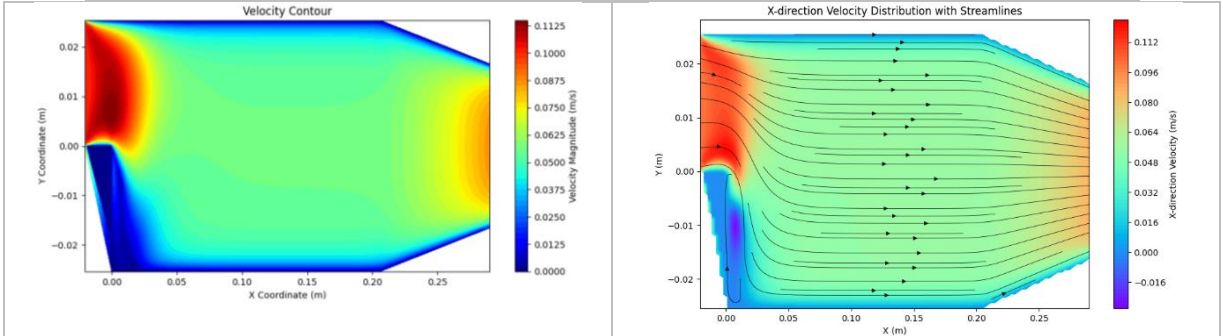

| (a) | (b) |
|---|---|

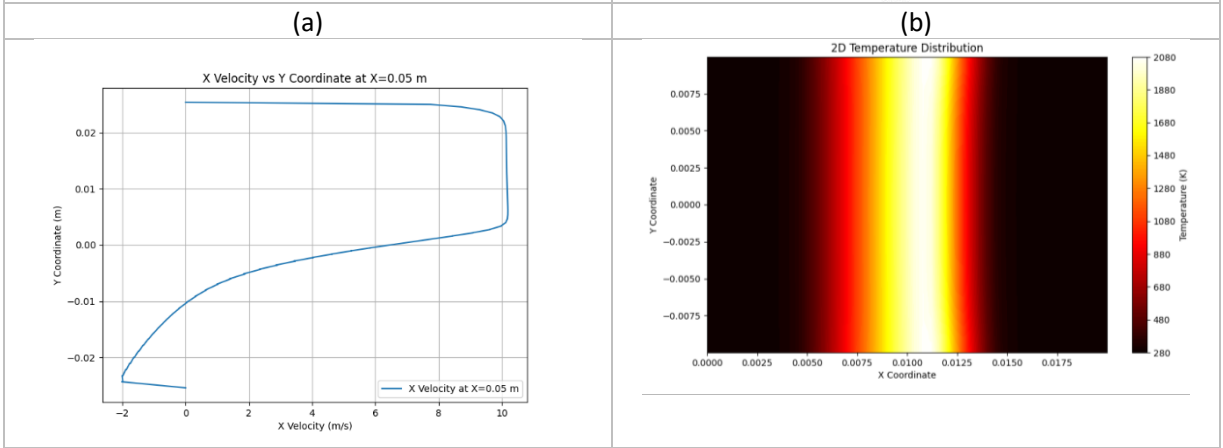

| (c) | (d) |
|---|---|

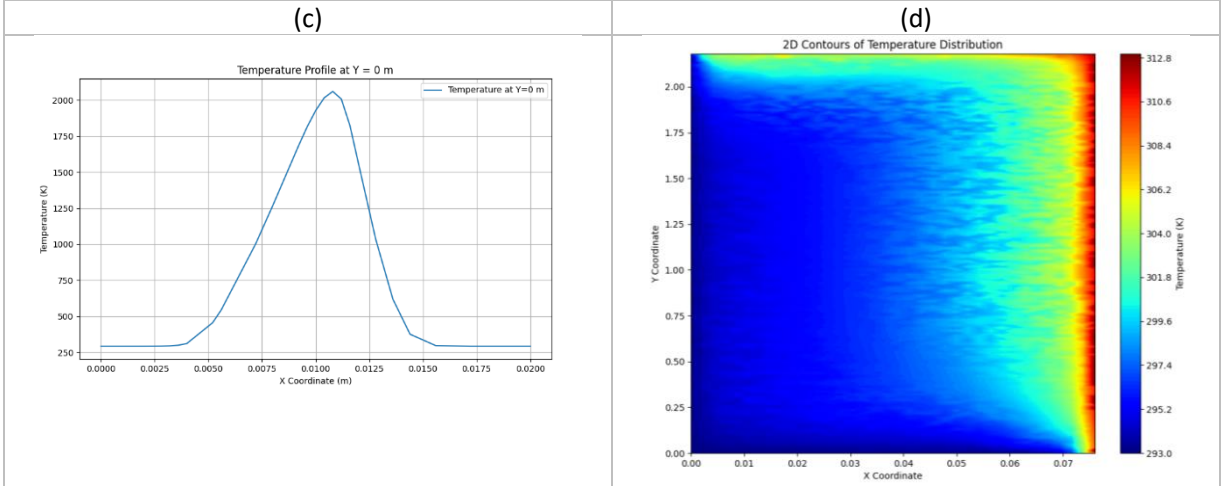

| (e) | (f) |
|---|---|

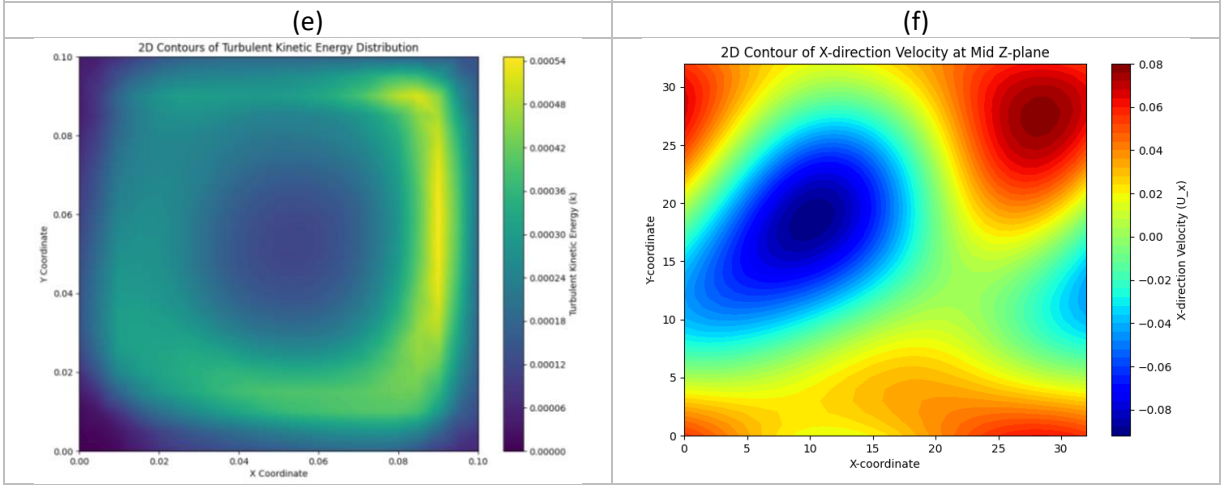

| (g) | (h) |
|---|---|

Figure 9: Postprocessing results of visualization using MetaOpenFOAM 2.0. (a) PitzDaily: Plot contour of U (b) PitzDaily: Plot contour of U with streamlines (c) PitzDaily: Plot profile of U (d) CounterFlowFlame: Plot contour of T (e) CounterFlowFlame: Plot profile of T (f) Cavity: Plot contour of TKE (g) BuoyantCavity: Plot contour of T (h) HIT: Plot X-velocity

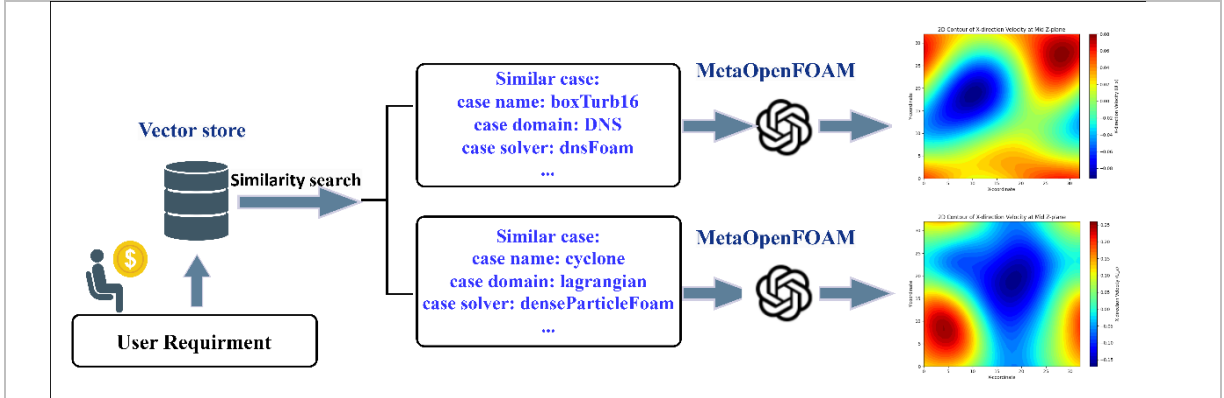

Figure 10: Different matches of RAG in MetaOpenFOAM (HIT case as example)

### A.4 Matching with low similarity

Although most of the simulation tasks in the benchmarking for CFD simulation and postprocessing can be found in OpenFOAM tutorials, there are instances where the system may match a case with low similarity. For example, in the HIT case, as shown in Figure 10, the system might be affected by randomness and match two different cases: boxTurb16, which is the most similar case in the database, and cyclone, which has low similarity to the CFD task. However, even when cyclone is matched, the CFD task named HIT still has a relatively high probability of passing the test. For instance, in the HIT test, there is a 50% chance of matching cyclone, but the actual pass rate reaches 75%. This demonstrates that MetaOpenFOAM has a certain capability to complete CFD simulation and post-processing tasks, even when matching cases with lower similarity.

### A.5 LLM-assisted verification

In this appendix, one case was selected from the postprocessing tasks of visualization and extraction to illustrate the role of LLM-assisted verification.

A. LLM-assisted verification for postprocessing in extraction

For example, when performing the task, *PitzDaily: Extract max Yplus*, a snippet of the generated Python script is as follows:

```python
# Iterate over all cells to find the
maximum yPlus value
for i in
range(output.GetNumberOfCells()):
    cell = output.GetCell(i)
    cell_data = output.GetCellData()
    yPlus_array =
cell_data.GetArray('yPlus')

    if yPlus_array:
        for j in
range(cell.GetNumberOfPoints()):
            yPlus_value =
yPlus_array.GetValue(j)
            if yPlus_value >
max_yPlus:
                max_yPlus =
yPlus_value
```

The output of this script is:

```
{"postprocessing_var": 0.0}
```

From a physical standpoint, it is clear that Yplus cannot be zero, which is obviously an error. Consequently, the feedback provided by the Reviewer is:

```
{
    "status": "failure",
    "Problem_description": " The
maximum `yPlus` value of 0.0 is not
within a reasonable range for wall-
bounded flows, typically aiming for
`yPlus` values between 30 and 300 for
wall functions or below 1 for
resolving the viscous sublayer.",
```

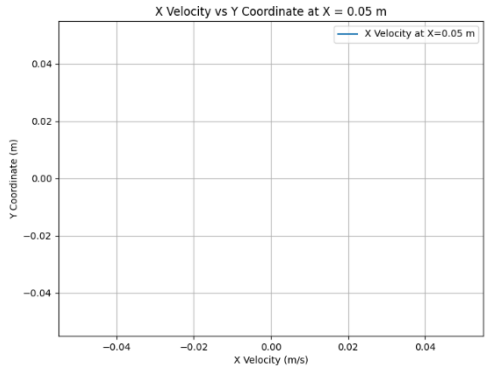

(a)

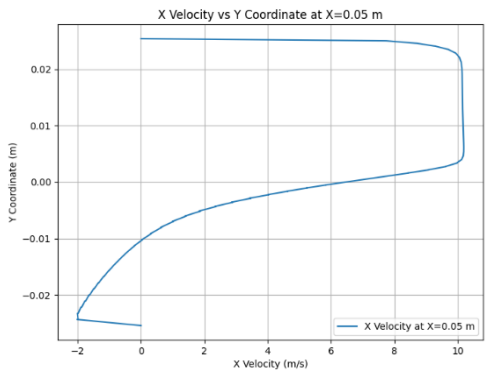

(b)

Figure 11: The Effect of LLM-Assisted Verification on the *PitzDaily: Plot profile of U* case (a) Before verification and (b) After verification and refinement

```
    "files_to_modify":
["postprocessing_python.py"]
}
```

Subsequently, the InputWriter modifies the postprocessing Python script (postprocessing_python.py) to change the method of extracting the Yplus array from *output.GetCellData()* to *output.GetPointData()*, as follows:

```
# Get the yPlus array from point data
(instead of cell data)
yPlus_array =
output.GetPointData().GetArray('yPlus
')
# Iterate over the yPlus_array to
find the maximum yPlus value
if yPlus_array:
    for i in
range(yPlus_array.GetNumberOfTuples()
):
        yPlus_value =
yPlus_array.GetValue(i)
        if yPlus_value > max_yPlus:
            max_yPlus = yPlus_value
```

The output of this modified script is:

```
{"postprocessing_var":
19.62164878845215}
```

This value is then deemed reasonable by the Reviewer.

B. LLM-assisted verification for postprocessing in visualization

After executing the Python script, an image as shown in Figure 11 (a) was generated, which only displayed the axis labels. Upon reviewing the image, the Reviewer provided feedback, identifying an error and describing the associated issue:

```
{
    "status": "failure",
    "Problem_description": "The plot
appears to be empty, indicating that
no data points were extracted or
plotted. This could be due to the X
coordinate not matching any points in
the dataset or an issue with reading
the VTK file.",
    "files_to_modify":
["postprocessing_python.py"]
}
```

In response to the problem description, the InputWriter modified the postprocessing Python script (postprocessing_python.py) by adding a debug check:

```
# Check if any data points were found
if len(y_coords) == 0:
    print("No data points found at X
= 0.05 m. Please check the VTK file
and the X coordinate.")
else:
```

As a result, the output "*No data points found at X = 0.05 m. Please check the VTK file and the X coordinate.*" was printed, indicating that no points were found at X = 0.05 m. Based on this debug output, the InputWriter made the following correction to the code:

```
# Loop through each point in the
dataset
for i in
range(points.GetNumberOfPoints()):
    x, y, z = points.GetPoint(i)
```

```
    if np.isclose(x, x_target, 
atol=1e-5):  # Check if the point is 
at the desired X coordinate
        y_coords.append(y)
        velocity = 
velocity_array.GetTuple(i)
        x_velocity.append(velocity[0]
)  # Extract the X component of the 
velocity
```

It was modified as:

```
# Loop through each point in the 
dataset
for i in 
range(points.GetNumberOfPoints()):
    x, y, z = points.GetPoint(i)
    if np.isclose(x, x_target, 
atol=1e-3):  # Adjusted tolerance to 
1e-3
        y_coords.append(y)
        velocity = 
velocity_array.GetTuple(i)
        x_velocity.append(velocity[0]
)  # Extract the X component of the 
velocity
```

The threshold was adjusted from *atol=1e-5* to *atol=1e-3*, leading to the correct result shown in Figure 11 (b).

These corrections in both the extraction and visualization tasks during actual execution demonstrate the effectiveness of LLM-assisted verification.